\documentclass[runningheads]{llncs}
\usepackage{subcaption}
\usepackage{floatrow}
\usepackage{booktabs}
\usepackage{multirow}
\usepackage{graphicx}
\usepackage[svgnames,table]{xcolor}
\definecolor{mygray}{gray}{0.95}
\usepackage{tikz}
\usepackage{comment}
\usepackage[hyperfootnotes=false]{hyperref}
\usepackage{amsmath,amssymb} 
\usepackage{color}
\usepackage{engord}

\newcommand{\figref}[1]{Figure~\ref{#1}}
\newcommand{\tabref}[1]{Table~\ref{#1}}
\newcommand{\secref}[1]{Section~\ref{#1}}

\usepackage[accsupp]{axessibility}  

\usepackage{marvosym}
\newcommand\blfootnote[1]{%
  \begingroup
  \renewcommand\thefootnote{}\footnote{#1}%
  \addtocounter{footnote}{-1}%
  \endgroup
}

\def\ie{\emph{i.e.}}

\def\vs{\emph{v.s.~}}

\begin{document}
\pagestyle{headings}
\mainmatter
\def\ECCVSubNumber{2652}  

\title{NSNet: Non-saliency Suppression Sampler \\ for Efficient Video Recognition} 

\titlerunning{NSNet: Non-saliency Suppression Sampler for Efficient Video Recognition}
%
\author{
{Boyang Xia$^{1,2}$\textsuperscript{*}} \and
Wenhao Wu$^{3,4}$\textsuperscript{*\Letter}\and
Haoran Wang$^{4}$\and 
Rui Su$^{5}$ \and Dongliang He$^{4}$ \and \\ Haosen Yang$^{6}$  \and Xiaoran Fan$^{2}$ \and Wanli Ouyang$^{5,3}$ \\
} 
\authorrunning{Boyang Xia$^*$ and Wenhao Wu$^*$\textsuperscript{\Letter} et al.}
%

\institute{Key Lab of Intelligent Information Processing of Chinese Academy of Sciences (CAS), Institute of Computing Technology, CAS, Beijing, China
\and  
University of Chinese Academy of Sciences, Beijing, China
\and 
SenseTime Computer Vision Group, The University of Sydney, Sydney, Australia
\and
Baidu Inc., Beijing, China
\and Shanghai AI Laboratory, Shanghai China \and Harbin Institute of Technology, Harbin, China
}

\maketitle

\begin{abstract}
It is challenging for artificial intelligence systems to achieve accurate video recognition under the scenario of low computation costs. Adaptive inference based efficient video recognition methods typically preview videos and focus on salient parts to reduce computation costs. 
Most existing works focus on complex networks learning with video classification based objectives. 
Taking all frames as positive samples, few of them pay attention to the discrimination between positive samples (salient frames) and negative samples (non-salient frames) in supervisions. To fill this gap, in this paper, we propose a novel \textbf{Non-saliency Suppression Network (NSNet)}, which 
effectively suppresses the responses of non-salient frames. 
Specifically, on the frame level, effective pseudo labels that can distinguish between salient and non-salient frames are generated to guide the frame saliency learning.
On the video level, a temporal attention module is learned under dual video-level supervisions on both the salient and the non-salient representations. Saliency measurements from both two levels are combined for exploitation of multi-granularity complementary information. Extensive experiments conducted on four well-known benchmarks
verify our NSNet not only achieves the state-of-the-art accuracy-efficiency trade-off but also present a significantly faster (2.4$\sim$4.3$\times$) practical inference speed than state-of-the-art methods. Our project page is at \url{https://lawrencexia2008.github.io/projects/nsnet}.
\blfootnote{*: Co-first authorship. This work was done when Boyang was an intern at Baidu. \\ \Letter: Corresponding author.}
\keywords{Video Recognition, Adaptive Inference, Temporal Sampling}
\end{abstract}


\begin{figure}[t]
      \centering \includegraphics[width=0.755\textwidth]{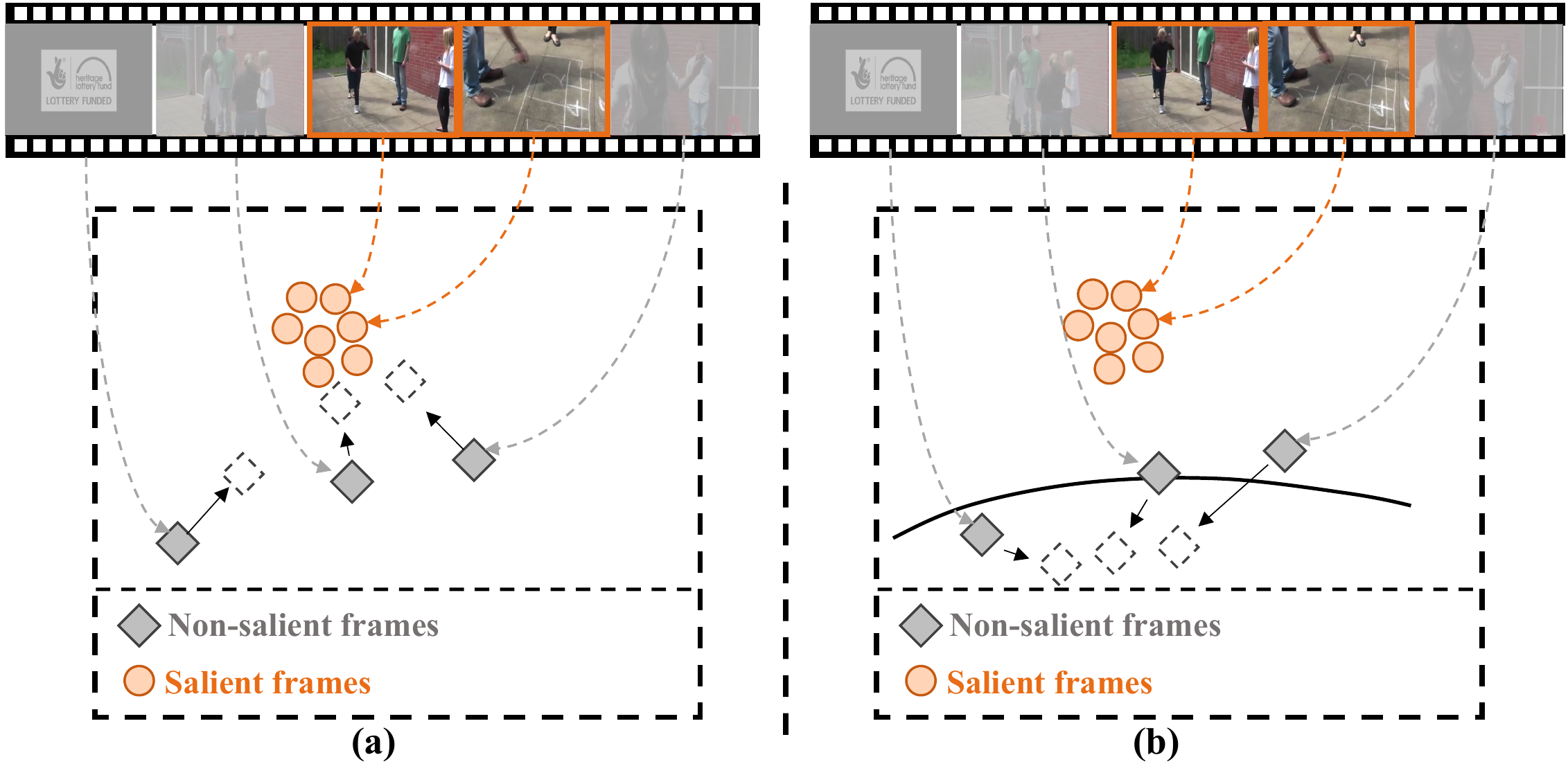}
      \caption{\textbf{A conceptual comparison between our proposed non-saliency suppression based sampler and existing approaches. } (a) Feature space learned by sampler networks 
      optimized by vanilla classification objectives with video labels. (b) Feature space learned by proposed NSNet.
       A video of \emph{Hopscotch} is used for illustration. The arrows indicate the moving directions of frame features during training process. In (a),
       the features of non-salient frames, the \engordnumber{1}, \engordnumber{2}, \engordnumber{5} frames are forced to cluster near the centroid of the category. 
       In contrast, our approach introduce a \emph{non-salient} category and those low saliency frames are labeled as \emph{non-salient} and function as negative samples against the video category during training. In this way, features of the \engordnumber{1}, \engordnumber{2}, \engordnumber{5} frames in (b) are pushed away to form another cluster of \emph{non-salient} category.
      }
      \label{fig:figure1}
\end{figure}
\section{Introduction}
\label{intro}
The prevalence of digital devices exponentially increases the data amount of video content. Meanwhile, the proliferation of videos poses great challenges for existing video analysis systems, and consequently draws more and more attention from the research community. 
Thanks to the renaissance of deep neural networks \cite{resnet,haoran_ijcai,gfnet}, a surge of progress has been made to promote the development of advanced video understanding techniques \cite{surui_2019_CVPR,surui_2021_ICCV,bcnet,wu2021weakly,AKnet,wang2020symbiotic,liu2019multi,huang2018toward}. 
Although achieving promising performance on some benchmarks \cite{kay2017kinetics,caba2015activitynet} with supervised learning or unsupervised learning~\cite{ASCNet,mamico}, many of them apply computationally heavy networks, which hinders their deployment to the practical applications, such as autonomous driving and personalized recommendations. Accordingly, building an efficient video understanding system is a crucial step towards widespread deployment in the real world.



To achieve efficient video recognition,
a rich line of studies have been proposed, which roughly fall into two paradigms: \romannumeral1) 
\textbf{lightweight architecture based methods} and \romannumeral2) \textbf{adaptive inference based methods}. The first category of approaches \cite{tsm,mvf} devote to reducing the computational cost via designing lightweight networks. By contrast, another series of works propose to achieve efficient recognition by leveraging adaptive inference strategy to flexibly allocate resources according to the the saliency of frames. Specifically, the adaptive inference mechanism has been applied on multiple dimensions of video, including temporal sampling \cite{scsampler19}, spatial sampling \cite{adafocus,arnet}, \emph{etc.} Compared to the former, the adaptive inference based methods is easier to be incorporated into the existing advanced recognition backbones. For example, in the pipeline of a adaptive temporal sampling method, a sampler network is first trained based on lightweight feature to sample key frames, and then an off-the-shell computation-consuming recognizer is evoked on sampled frames for final recognition. 

Semantic saliency of each frame is the fundamental basis for the adaptive inference based methods \cite{scsampler19,adamml}.  Nonetheless, it is difficult to obtain explicit supervision of frame-level saliency in the general setting of video recognition. Therefore, existing methods are mainly based on either reinforce learning (RL) or attention mechanism~\cite{adaframe,marl,arnet,smart2020}, 
where the agent (\textit{resp.}, attention module) is optimized to take actions (\textit{resp.}, attend) to those salient frames by classification objective based rewards (\textit{resp.}, loss).
In this way, the sampler is trained to determine the salient frames by using all frames as the positive samples of the corresponding video category, as shown in \figref{fig:figure1}. Due to the lack of negative samples in training, it is hard for the sampler to accurately determine the non-salient frames from the salient ones within one video, which may easily overestimate the saliency of the non-salient frames. As a result, it is reasonable to introduce the negative samples for the frame-level video category classification objective based learning, which helps suppress the response of the non-salient frames to the video category during the sampling process.

To this end, we propose a novel \textbf{Non-saliency Suppression (NS) mechanism}, to provide negative-sample-aware supervision for saliency measurement, which can effectivly suppress the response of non-salient frames in the adaptive inference based framework. 
Specifically, the key principle is that the salient frames should belong to the corresponding video category while the non-salient frames should fall into a special category, which is distinguishable from all video categories. We term this special category as \textit{non-salient} category, as shown in \figref{fig:figure1}. As the video categories are the only annotation we can use during training, in order to guarantee high-quality negative samples (\textit{i.e.,} the non-salient frames) in the frame-level saliency learning, we propose a Frame Scrutinize Module (FSM) to generate frame-level pseudo labels for supervision. 
By doing so, the salient frames can be effectively distinguished from the non-salient frames. In addition to the frame-level supervision, we then propose a temporal attention module named Video Glimpse Module (VGM) to compensate for high-level information of video events by using video-level supervision. In order to introduce NS mechanism on video level, we first formulate a video representation as a linear combination of two components: the representation of the salient parts and the representation of the non-salient parts of a video. Following the aforementioned principle, we then assign the label of that salient representation as current video label, and the label of non-salient representation as the special \emph{non-salient} category.
Overall, our contributions are three-folds:
\begin{itemize}
    \item We introduce the \textbf{Non-saliency Suppression mechanism} for suppressing the responses of non-salient frames, which considerably improve the discrimination power of the temporally sampled video representations without increment of computation overhead.
    \item We propose an discriminative and flexible multi-granularity frame sampling framework \textbf{N}on-saliency \textbf{S}uppression \textbf{Net}work (\textbf{NSNet}), which leverages supervisions from both video level and frame level to measure frame saliency. We design two specific schemes for realizing Non-saliency Suppression mechanism on the two granularities. 
    \item Extensive experiments are conducted with multiple backbone architectures on four well-known benchmarks, \ie, ActivityNet, FCVID, Mini-Kinetics and UCF101, which show that our NSNet achieves superior performance over existing state-of-the-art methods with limited computational costs. 
\end{itemize}

\section{Related Work}
\label{relatedwork}
\subsubsection{Video Recognition.} Video Recognition has made significant progress in past decade for successful application of neural networks including 2D CNNs~\cite{tsn}, 3D CNNs~\cite{i3d} and Transformers \cite{timesformer,Wu2022TransferringTK}. 
Although decent results are achieved by powerful spatiotemporal networks,
it is still challenging for applying video recognition in resource-constraint scenarios for its superfluous computational complexity. TSM \cite{tsm}, TEA~\cite{tea2020}, MVFNet~\cite{mvf}, \emph{etc.}, try to realize temporal modelling with pure 2D CNNs to improve efficiency by shifting operations, motion based channel selection and multi views fusion. While P3D \cite{p3d}, S3D \cite{s3d}, R(2+1)D \cite{r2plus1d}, SlowFast \cite{slowfast}, Ada3D \cite{ada3d}, DSANet \cite{dsanet} are proposed to improve the efficiency by decomposing 3D convolution or designing hybrid 2D-3D frameworks. Different from these approaches, we seek to achieve efficient video recognition by adaptively sampling salient frames and recognize selectively on a per-sample basis.

\noindent\textbf{Adaptive Inference.} The core idea of adaptive inference is to dynamically allocate computational resources (network layers, parameters, \emph{etc.}) conditioned on the input to improve the trade-off between performance and cost \cite{dynamic_survey,wu2020dynamic}.
For video analysis, adaptive inference are realized in several perspectives including temporal sampling, resolution, sub-networks and modality. FastForward \cite{fastforward}, FrameGlimpse \cite{frameglimpse}, AdaFrame \cite{adaframe}, MARL \cite{marl} and OCSampler \cite{ocsampler} model temporal sampling as a decision-making process, which is optimized by policy gradients or differentiable alternatives.  ListenToLook \cite{listentolook}, SMART \cite{smart2020}, TSQNet~\cite{tsqnet} design temporal frame samplers based on attention mechanism. Besides temporal sampling, AdaFocus series \cite{adafocus,adafocusv2} samples salient patches for each frame to reduce spatial redundancy. AR-Net \cite{arnet}, LiteEval \cite{liteeval}, AdaMML \cite{adamml} strategically allocate higher resolution, more powerful sub-networks, or more expensive modality to more informative frames, respectively. 

The most closely related work to ours is an adaptive temporal sampling method, SCSampler \cite{scsampler19}, which proposes a frame-level classification task with video label and measure saliency based on classification confidence. However, there exist substantial differences between SCSampler and the proposed NSNet. SCSampler assigns each frame with the video label, while we argue that only the salient ones belong to video category, other ones should be labeled as a special category distinguishable from all semantic categories. Besides, we also consider to measure frame saliency with video level supervisions to enable context-aware saliency measurements, which is overlooked by SCSampler.
Compared with SCSampler, our NSNet achieves significantly superior performance with much less computational overhead.
\section{Approach}
\label{sec:approach}
\begin{figure*}[t]
      \centering
      \includegraphics[width=0.90\textwidth]{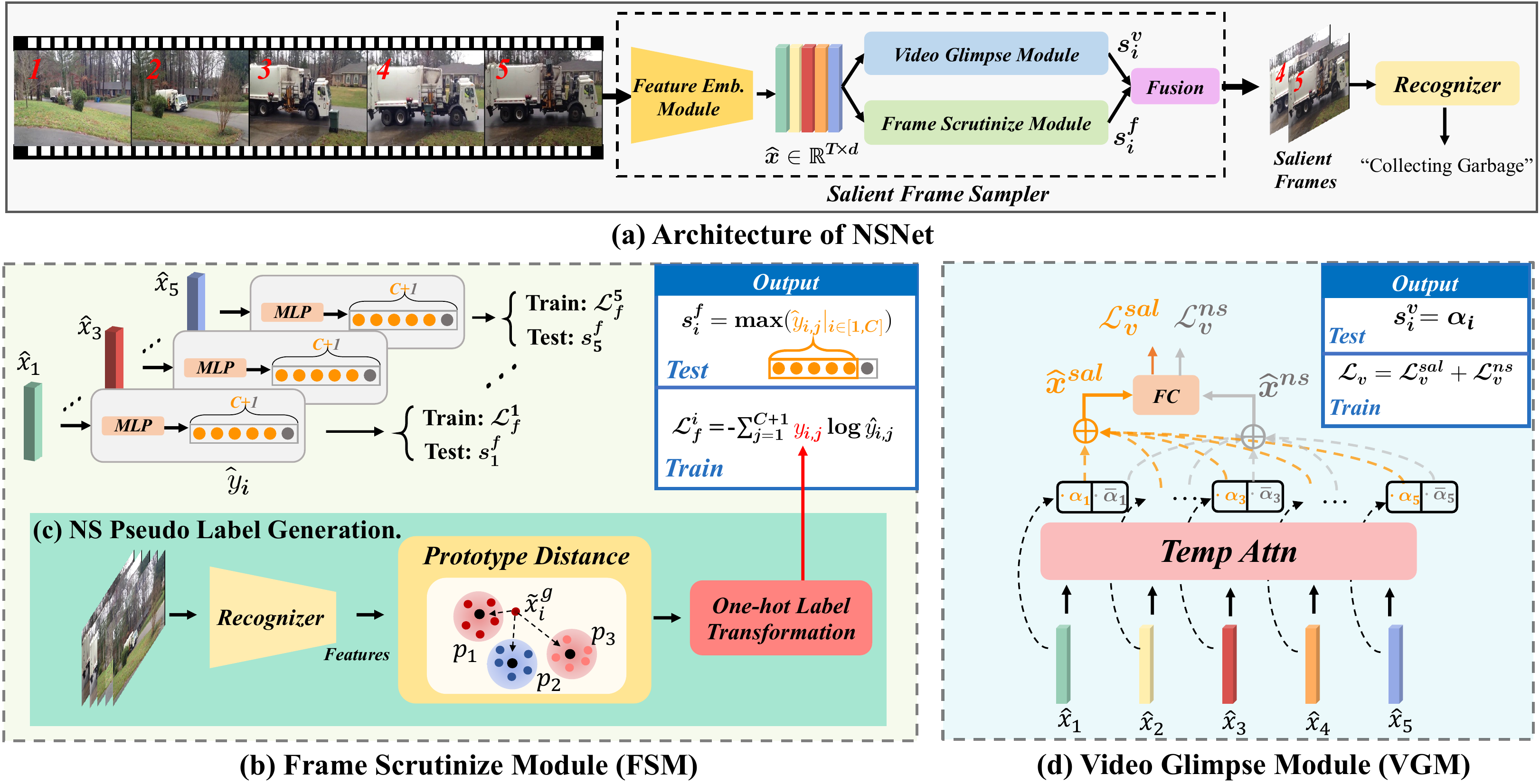}
      \caption{\textbf{An overview of architecture of NSNet.} 
    (a) shows the whole architecture. (b) shows the Frame Scrutinize Module (FSM), which estimates the saliency of each frame by the prediction confidence in frame-level classification. (c) shows the proposed Non-saliency Suppression (NS) frame-level pseudo label generation strategy based on the distance between each frame and video category prototypes. (d) shows the Video Glimpse module (VGM), which measures the saliency of each frame using temporal attentions in video-level classification.
      }
      \label{fig:arch}
      
\end{figure*} 

\subsection{Problem Definition}\label{subsec:Problem Formulation}
Let $V=\{x_i\}_{i=1}^T$ denote a video of $T$ frames. Firstly, our NSNet is designed to select $K$ most salient frames from all $T$ frames. Then, the $K$ salient frames are fed into a recognizer model, the predictions of which are aggregated to yield the final video-level prediction. The workflow of our system is illustrated in  \figref{fig:arch}a. Note that we use an off-the-shell model as our recognizer, and therefore, the problem can be formulated as how to effectively sample the most salient frames from the input frames. In the following sections, we introduce the process of salient frames selection for our NSNet.
\subsection{The Overview of NSNet}
In this section, we elaborate on the proposed \textbf{N}on-saliency \textbf{S}uppression \textbf{Net}work (NSNet), which mainly consists of three components: a \textbf{Feature Embedding module (FEM)}, a \textbf{Frame Scrutinize module (FSM)}, and a \textbf{Video Glimpse module (VGM)}. The feature embedding module generates feature embedding from the input video frames. The FSM (see \figref{fig:arch}b) measures saliency of each frame by predicting the saliency confidence scores in frame-level classification, and the VGM (see \figref{fig:arch}d) models saliency of frames from the temporal attention weights used to aggregate attention-based feature for video-level classification. To alleviate the lack of negative samples, we further apply NS mechanism in the FSM and VGM in different ways. For each of these two modules, a \emph{non-salient} category is attached onto the frame-level and the video-level supervisions, respectively, resulting in a total of $C+1$ categories for each supervision, where $C$ is the total number of the original video categories. In the FSM, a Non-saliency Suppression frame-level pseudo label generation strategy is proposed to separate the negative samples from the truly salient frames for frame-level saliency learning. In the VGM, a Non-saliency Suppression loss is proposed to impose an extra constraint of non-salient representations of videos besides original classification objectives.

\subsection{Feature Embedding Module (FEM)}
\label{sec:embed}
Here we encode the input video frames $\{x_i\}_{i=1}^T$ into the robust feature sequence $\{\hat{x}_i\}_{i=1}^T$, which is then used by our FSM and VGM.
Specifically, we first use the off-the-shelf lightweight feature extractor, \textit{e.g.,} MobileNet, EfficientNet, \emph{etc}, to take the video frames as input and extract features for all frames. In order to allow message passing among features for all frames, we then apply a transformer encoder~\cite{transformer} on top of these features and output the feature sequence $\{\hat{x}_i\}_{i=1}^T$.
\subsection{Frame Scrutinize Module (FSM)} \label{fsmodule}

In our FSM, we first generate frame-level Non-saliency Supression (NS) pseudo labels, and then use them as supervision to train our FSM to perform frame-level saliency classification and produce saliency scores. Details of our FSM are provided as follows.

\noindent \textbf{Non-saliency Suppression Pseudo Label Generation.}
Here we denote the label of a video of the $c$-th category as a $C$-dimension one-hot vector $y_v\in \mathbb{R}^{C}$, where $y_{v,c}=1$ and $y_{v,m}=0|_{m\in[1,C], m\neq c}$. 
To distinguish between the salient frames and non-salient frames in frame labels, we then introduce a \textbf{guiding saliency score} $g_i$, which is obtained from the recognizer, \emph{i.e.}, the one we used for final recognition as described in Section~\ref{subsec:Problem Formulation} (Figure \ref{fig:arch}a). Although the classification response produced by the pre-trained model (\emph{i.e.,} recognizer) is widely used for pseudo labeling in weakly-supervised learning \cite{wsod_oicr,bgmodel,wsss}, we propose a prototype-based strategy for more robust frame level pseudo label generation.
According to \cite{proto_network,zhang2020discriminability}, a sample could be more representative when it is closer to the centroid in feature space. As a result, we use distances of the feature for the $i$-th frame from the prototype features of all categories to obtain  $g_i$.
Specifically, we first use the recognizer to extract features and confidence scores on ground-truth category for all frames for each video in training set. Then the prototype feature of each category is then calculated by averaging all video features in that category. Here each video feature is obtained by applying average pooling on the features of the top-K frames based on the predicted confidence scores (see our Appendix \ref{appendix:train_details} for more details)
The guiding saliency score $g_i$ for the $i$-th frame is as follows:
\begin{equation}
    g_{i} = \frac{e^{\phi(\Tilde{x}^{g}_i,p_c)}}{\sum_{j=1}^{C} e^{\phi(\Tilde{x}^{g}_i,p_j)}}, 
\end{equation}
where $\operatorname{\phi}$ is a distance function measuring the similarity of two feature vectors, \emph{e.g.,} Euclidean Distance, $\Tilde{x}^{g}_i$ is the feature for the $i$-th frame extracted by the recognizer, $p_j$ and $p_c$ are the prototype features of the $j$-th category and the ground truth category, respectively.
Finally, we use $g_i$ to generate the NS pseudo label $y^{ns}_i= [\ g_iy_{v,1},\ g_iy_{v,2},\ ...\  g_iy_{v,C},1-g_i]\in \mathbb{R}^{C+1}$.

\noindent \textbf{Frame-level Saliency Classification.}
After generating the NS pseudo labels, we then use them to train our FSM to perform frame-level saliency classification over the feature sequence $\{\hat{x}_i\}_{i=1}^T$. Mathematically, the frame-level classification objective is defined as follows:
\begin{equation}
    \mathcal{L}_{f}={-\sum_{i=1}^{T}\sum_{j=1}^{C+1}y^{ns}_{i,j}} \log(\hat{y}^{ns}_{i,j}),
\end{equation}
where $y_{i,j}^{ns}$ is the element of the $j$-th category in $y_i^{ns}$, and $\hat{y}^{ns}_{i,j}$ is the classification prediction. It is noteworthy that $y_i^{ns}$ is a soft one-hot target, the cross entropy loss of which is similar to label smooth \cite{labelsmooth}.
During inference, the frames with very high response to any one of $C$ categories are identified as salient frames. To this end, the maximum confidence across $C$ semantic categories (except the $C+1$-th category) of classification score after softmax normalization is used for saliency measurement. We then apply additional softmax normalization along the time axis to obtain final saliency score $s^f_i$.

\subsection{Video Glimpse Module (VGM)}\label{vgmodule}
In our VGM, we first generate attention weights $\alpha_i =  \operatorname{TempAttn}(\hat{x}_i)$ for the features of all observed frames, where $\operatorname{TempAttn}(\cdot)$ is implemented by a fully-connected layer followed by a L1 normalization layer, which is used to rescale attention weights to $[0,1]$ range. The features of all observed frames are then aggregated with the attention weights to generate the video salient representation $\hat{x}_{v}^{sal}=  \sum_{i=1}^T {\alpha}_{i} \hat{x}_i$.
To perform video-level classification, the salient representation of the video $\hat{x}_{v}^{sal}$ is fed to a fully-connected layer to compute the cross-entropy loss with video label. In order to guide our VGM to separate negative samples (non-salient frames) from positive samples (salient frames) of current video category, we propose a Non-saliency Suppression (NS) loss to impose a constraint other than the regular classification objective. During inference, the attention weights are used as saliency scores $\{s^{v}_i\}_{i=1}^T$. The details of the NS loss are described next.

\noindent\textbf{Non-saliency Suppression Loss.}
It is obvious that all videos contains both salient and non-salient frames for a specific video category.
Therefore, it is natural that a holistic video representation $\hat{x}_{v}$ can be formulated as a linear combination of the salient representation $\hat{x}_{v}^{sal}$ and the non-salient representation $\hat{x}_{v}^{ns}$ \cite{bgmodel}, \textit{i.e.,} $\hat{x}_{v} = \hat{x}_{v}^{sal} + \gamma \hat{x}_{v}^{ns}$. The non-salient representation $\hat{x}_{v}^{ns}$ can be obtained as follows. We first compute the complementary part of attention weights $\overline{\alpha}_i = \frac{1}{T}(1-\alpha_{i})$, and then
aggregate the feature sequence with $\overline{\alpha}_i$ and produce the non-salient representation $\hat{x}_{v}^{ns}=\sum_{i=1}^T \overline{\alpha}_{i} \hat{x}_i.$
In this way, a video can be regarded as a positive sample of both its ground truth category and \textit{non-salient} category to different proportions, at the same time. 
Both $\hat{x}_{v}^{sal}$ and $\hat{x}_{v}^{ns}$ will be fed into the classification fully-connected layer to get different predictions $\hat{y}^{sal}_{v}$ and $\hat{y}^{ns}_{v}$. 
Then we defines labels for both $\hat{y}^{sal}_{v}$ and $\hat{y}^{ns}_{v}$:
$y^{ns}_{v}=[0,0,...,0,1]\in \mathbb{R}^{C+1}$,\, $y^{sal}_{v}=[y_{v,1},y_{v,2},...y_{v,C},0] \in \mathbb{R}^{C+1}$, where $y_v$ is the original video label.
The cross-entropy loss between $\hat{y}^{sal}_{v}$ and $y^{sal}_{v}$ is the original classification loss $\mathcal{L}_{cls}$, and the one between $\hat{y}^{ns}_{v}$ and $y^{ns}_{v}$ is the NS loss $\mathcal{L}_{ns}$.
Consequently, the objective function of this module is defined as follows, where $\gamma$ is the weight of $\mathcal{L}_{ns}$.
\begin{equation}
    \mathcal{L}_v = \mathcal{L}_{cls} + \gamma \mathcal{L}_{ns},
\end{equation}
\label{v_loss} 
\subsection{Learning Objectives}
The overall objective function of our NSNet is formulated as follows:
\begin{equation}
    \mathcal{L} =  \mathcal{L}_v + \mathcal{L}_f,
\end{equation}\label{total_loss}
where $\mathcal{L}_v$ and $\mathcal{L}_f$ denote the loss function of the VGM and the FSM, respectively. This objective not only drives model to conduct discriminative saliency measuring according to video semantics and frame discrepancy, but also facilitates information exchange between the video context and parts in shared feature encoding. 
\section{Experiments}
\subsection{Datasets and Evaluation Metrics}
We evaluate our method on four large-scale video recognition benchmarks, \emph{i.e.,} ActivityNet, FCVID, Mini-Kinetics and UCF101.
ActivityNet~\cite{caba2015activitynet} contains 19994 videos of 200 categories of most popular actions in daily life.
FCVID~\cite{fcvid} contains 91,223 videos collected from YouTube and divided into 239 classes covering most common events, objects, and scenes in our daily lives.
Mini-Kinetics~\cite{arnet} is a subset of Kinetics~\cite{kay2017kinetics} presented by \cite{arnet}, including 200 categories of videos of Kinetics, with 121k videos for training and 10k videos for validation.
UCF101~\cite{ucf101} has 101 classes of actions and 13K videos with short duration (7.2sec).
Mean Average Precision (mAP) is used as the main evaluation metric for ActivityNet and FCVID, while Top-1 accuracy is used for Mini-Kinetics and UCf-101 following previous works.  
We also report the computational cost (in FLOPs) to evaluate the efficiency of the proposed method. FLOPs of our method are composed of following parts:
$P_{total} = P_{rec} \times K + P_{fem}+P_{vgm}+P_{fsm},$
where $P_{rec}, P_{fem}, P_{vgm}, P_{fsm}$ represent the FLOPs of the 
recognizer, FEM, VGM and FSM, respectively. An example of FLOPs computation of our model with the setting in \tabref{tab:anet_sota:Res50} is: 4.109$\times$5 (ResNet-50 with 5 frames) + (0.320$\times$16+0.315) (MobileNetv2 with 16 frames+transformer) + 0.004 + 0.002 = 25.99(G).
\subsection{Implementation Details}\label{sec:details}
\textbf{Training.} 
Following previous works \cite{smart2020,adaframe}, we mainly use MobileNetv2 \cite{mobilenetv2} as the lightweight feature extractor in our FEM. Different high-capacity networks trained on target datasets are used as recognizers at the same time: ResNet family \cite{resnet} and Swin-Transformer family \cite{swintransformer}, \emph{etc.}
For the transformer encoder in our FEM, 2 encoder layers with 8 heads and learnable positional embedding are used.
The distance function $\phi$ in our FSM used for guiding saliency score is Euclidean Distance. The non-saliency suppression loss weight $\gamma$ 
is set to 0.2. 
See Appendix \ref{appendix:train_details}.

\noindent\textbf{Inference.}
\label{sec:infer}
We fuse the results of FSM and VGM to obtain the final saliency measurements. Score \emph{sum, max, mul} and index \emph{union, intersect, join} are considered for fusion. See Appendix \ref{appendix:test_details} for details.

\subsection{Main Results}
\label{main_results}
\noindent\textbf{Comparison with Simple Baselines.}
As shown in \tabref{tab:simple_baseline}, we compare our approach with multiple hand-crafted sampling methods on ActivityNet and UCF101 with ResNet-101 and ResNet-50 as recognizers (without TSN training strategy \cite{tsn}), respectively. The simple baselines include \textsc{Uniform}, \textsc{Random}, \textsc{Dense}, and \textsc{Top-K} sampling. For \textsc{uniform} and \textsc{random}, we uniformly and randomly sample 10 frames from all frames, respectively, while for \textsc{Top-K}, we sample top 10 frames with highest predicted confidence scores (\ie, the maximum confidence among all categories), from all frames. For our method, we first uniformly sample an observation number (100 for ActivityNet and 50 for UCF-101) of frames as the observation frames from the input videos, and then use our method to sample 5 frames from the observation frames. \textsc{Dense} sampling makes use of all frames for recognition. We observe that our NSNet outperforms all simple baselines by a large margin on both two datasets. In ActivityNet, our method relatively outperforms the competitive but heavy \textsc{Top-K} baseline by 2.4\% in terms of mAP with 11.7$\times$ less GFLOPs, which verifies the effectiveness of our sampler. In UCF-101, the videos are much shorter than those in ActivityNet (7sec \vs 119sec on average), which constructs a much more difficult setting for sampler. However, the top-1 accuracy of our NSNet still relatively exceeds that of the most competitive \textsc{Dense} baseline by 2.1\% with much less GFLOPs, which demonstrates NSNet can improve the video classification performance on trimmed videos. 
\begin{table}[t]
\centering
\caption{Comparison with several hand-crafted sampling strategies. ResNet-101 and ResNet-50 are adopted as the recognizers for ActivityNet and UCF-101, respectively. }
\label{tab:simple_baseline}
\setlength{\tabcolsep}{6.0pt}
\renewcommand{\arraystretch}{0.9}
\scalebox{0.85}{
\begin{tabular}{c|cc|cc}
\toprule
& \multicolumn{2}{c|}{ActivityNet} & \multicolumn{2}{c}{UCF101} \\ \cmidrule(lr){2-3} \cmidrule(lr){4-5}
    & mAP(\%)           & FLOPs         & Top-1(\%)      & FLOPs      \\
\midrule
\textsc{Uniform}  & 68.6          & 195.8G           & 75.9          & 61.7G      \\
\textsc{Random}   & 68.1          & 195.8G           & 75.7             & 61.7G          \\
\textsc{Dense}      & 69.0            & 930.8G          & 76.1          & 753.4G     \\
\textsc{Top-K}   & 72.5          & 930.8G          & 74.5          & 753.4G     \\ \midrule
Ours    & \textbf{74.9}          & \textbf{73.2G}           & \textbf{77.6}          & \textbf{37.6G} \\  
\bottomrule
\end{tabular}
\scalebox{0.85}{
}
}
\end{table}
\begin{table}[t]
\caption{Comparisons with SOTA efficient video recognition methods with ResNet50 as the main recognizer on the ActivityNet dataset. The backbones used for sampler and recognizer are reported. MBv2 denotes MobileNetv2.}
\label{tab:anet_sota:Res50}
\centering
\setlength{\tabcolsep}{8pt}
\scalebox{0.8}{
\begin{tabular}{ccccc}
\toprule
Method  & Backbone & mAP(\%)  & FLOPs\\
\midrule
SCSampler\cite{scsampler19}  & MBv2+Res50 & 72.9 & 42.0G \\
AR-Net~\cite{arnet}  & MBv2+ResNets      & 73.8 & 33.5G  \\
AdaMML~\cite{adamml}  & MBv2+Res50 & 73.9& 94.0G \\
VideoIQ~\cite{videoiq}  & MBv2+Res50      & 74.8 & 28.1G  \\
AdaFocus~\cite{adafocus}  & MBv2+Res50      & 75.0 & 26.6G  \\
Dynamic-STE~\cite{dynamicSTE} & Res18+Res50 & 75.9 & 30.5G \\
FrameExit~\cite{frameexit} & ResNet-50   & 76.1 & 26.1G  \\
\midrule
\textbf{Ours} & MBv2+Res50  & \textbf{76.8}& \textbf{26.0G} \\
\bottomrule
\end{tabular}
}
\end{table}
\begin{table}[thb]
\caption{Comparisons with SOTA video recognition methods with ResNet152 and more advanced networks as the recognizers on the ActivityNet dataset.}
\label{tab:anet_sota:Res152}
\centering
\scalebox{0.8}{
\setlength{\tabcolsep}{1pt}{
\begin{tabular}{ccccc}
\toprule
Method  & Recognizer & Pretrain & Top-1(\%)  & mAP(\%)\\
\hline
\multicolumn{5}{l}{\cellcolor{mygray}~~ResNet-152 w/ ImageNet} \\
P3D~\cite{p3d} & ResNet-152     & ImageNet   & 75.1 & 78.9 \\
RRA~\cite{rra} & ResNet-152     & ImageNet   & 78.8 & 83.4 \\
MARL~\cite{marl}  & ResNet-152     & ImageNet  & 79.8 & 83.8 \\
ListenToLook~\cite{listentolook}  & ResNet-152     & ImageNet &  80.3 & 84.2  \\
\textbf{Ours}   &  ResNet-152     & ImageNet   &  \textbf{80.7}   & \textbf{85.1} \\
\hline \multicolumn{5}{l}{\cellcolor{mygray}~~ResNet-152 w/ Kinetics} \\
SMART~\cite{smart2020}     & ResNet-152     & Kinetics   & - & 84.4  \\
\textbf{Ours}       & ResNet-152     & Kinetics   & \textbf{84.5}   & \textbf{88.7}\\ 
\hline \multicolumn{5}{l}{\cellcolor{mygray}~~More Advanced Networks w/ Kinetics} \\
DSN~\cite{dsn}  &  R(2+1)D-34     & Kinetics &  82.6 & 87.8 \\
Ada3D~\cite{ada3d}  &  SlowOnly-50     & Kinetics &  - & 84.0 \\
ListenToLook~\cite{listentolook}  & R(2+1)D-152     & Kinetics &  - & 89.9  \\
MARL~\cite{marl}  &  SEResNeXt-152     & Kinetics &  85.7 & 90.1 \\
Ours   &  Swin-B     & Kinetics   &  86.7   & 91.6\\
\textbf{Ours}   &  Swin-L     & Kinetics   &  \textbf{90.2}   & \textbf{94.3}\\
\bottomrule
\end{tabular}
}}
\end{table}

\begin{figure}[h]
\floatbox[{\capbeside\thisfloatsetup{capbesideposition={left,top},capbesidewidth=5.5cm,margins=raggedright}}]{figure}[\FBwidth]
{\caption{Comparison with state-of-the-art sampling methods on ActivityNet dataset. 
      Our proposed NSNet achieves better mAP with much fewer GLOPs (per video) than other methods. It is worth noting that we compare these methods with the same recognizer ResNet-101, under different computation budgets. The results are quoted from the published works~\cite{listentolook,arnet}.}~\label{fig:anet_res101}
      }
{\includegraphics[width=5.5cm]{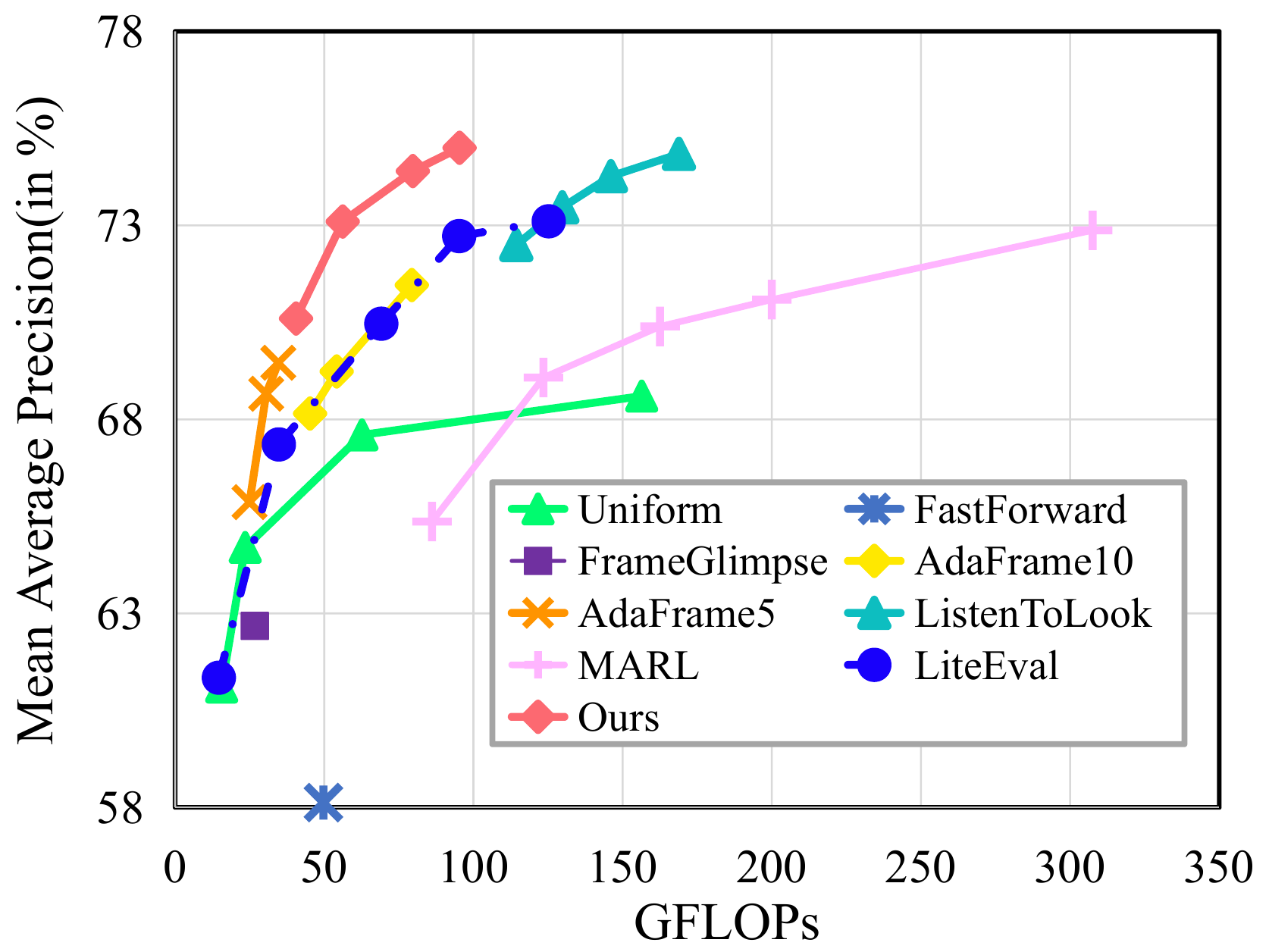}}
\end{figure}

\noindent\textbf{Comparison with SOTA on ActivityNet.}
We make a comprehensive comparison with recent state-of-the-art efficient video recognition methods on the ActivityNet dataset in Table \ref{tab:anet_sota:Res50}-\ref{tab:anet_sota:Res152}, and \figref{fig:anet_res101}. 
As shown in \tabref{tab:anet_sota:Res50}, we compare our NSNet with other state-of-the-art efficient approaches using ResNet-50 as the recognizer. NSNet consistently outperforms all existing methods including sampler-based and sampler-free approaches.
When compared with AR-Net \cite{arnet}, an adaptive resolution method, the mAP of our method improve by 3\% with much less computational cost (\textbf{26.0G} \vs 33.5G). Our NSNet also outperforms AdaMML \cite{adamml}, an adaptive modality approach, by 2.9\% in terms of mAP while having 3.6$\times$ less FLOPs. In addition, sampler-free approaches often have relatively low FLOPs, because they do not need sampling process.
However, although our NSNet has extra computational on the sampling process, it can achieve higher accuracy than FrameExit \cite{frameexit}(\textbf{76.8\%} \vs 76.1\%), a competing sampler-free framework, with comparable FLOPs, which demonstrates our sampler greatly improve the discrimination power of the video representation. 
In \tabref{tab:anet_sota:Res152}, we show the results of the SOTAs using ResNet-152 and more advanced backbones on the ActivityNet dataset. We can see that our NSNet outperforms the sampler-free methods P3D \cite{p3d}, RRA \cite{rra} by at least 1.7\% in terms of mAP. When compared with competing sampler-based methods (MARL \cite{marl}, ListenToLook \cite{listentolook}, SMART \cite{smart2020}), our method still shows superiority over them. Besides, when using a more advanced transformer-based network Swin-Transformer \cite{swintransformer} as the recognizer in our NSNet, the mAP can be further promoted to 94.3\%, which is the highest performance on ActivityNet to our best knowledge.
 
\figref{fig:anet_res101} illustrates the GFLOPs-mAP curve on the ActivityNet dataset. On this curve, the observation number $T$ is set to 50, while the number of sampled frames $K$ is tuned up as FLOPs budget increases. 
Following previous works \cite{adaframe,liteeval,listentolook,marl}, we use ResNet-101 without TSN-style training as the recognizer.
Our NSNet presents significant accuracy improvement with much lower GFLOPs than other methods.  



\begin{table}[t]
\caption{Comparison with previous methods on FCVID and Mini-Kinetics. Our NSNet consistently outperforms state-of-the-art  in terms of accuracy and efficiency using ResNet-50 as the recognizer.}
\label{tab:sota_fcvid_minik}
\centering
\setlength{\tabcolsep}{8pt}
\scalebox{0.8}{
\begin{tabular}{c|cc|cc}
\toprule
\multirow{2}{*}{Methods} & \multicolumn{2}{c|}{FCVID} & \multicolumn{2}{c}{Mini-Kinetics} \\ \cmidrule(lr){2-3}\cmidrule(lr){4-5} & mAP(\%) & FLOPs & Top-1(\%) & FLOPs\\
\midrule 
LiteEval~\cite{liteeval} & 80.0 & 94.3G & 61.0 & 99.0G \\
AdaFrame~\cite{adaframe} & 80.2 & 75.1G & - & - \\
SCSampler~\cite{scsampler19} & 81.0 & 42.0G & 70.8 & 42.0G \\
AR-Net~\cite{arnet} & 81.3 & 35.1G   & 71.7 & 32.0G \\
AdaFuse~\cite{adafuse} & 81.6 & 45.0G & 72.3 & 23.0G \\
SMART~\cite{smart2020} & 82.1 & - & - & - \\
VideoIQ~\cite{videoiq} & 82.7 & 27.0G & 72.3 & 20.4G \\
Dynamic-STE~\cite{dynamicSTE} & - & - & 72.7 & 18.3G \\
FrameExit~\cite{frameexit} & - & - & 72.8 & 19.7G \\
AdaFocus~\cite{adafocus} & 83.4 & 26.6G & 72.9 & 38.6G \\ \midrule
\textbf{Ours} & \textbf{83.9} & \textbf{26.0G} & \textbf{73.6} & \textbf{18.1G} \\     
\bottomrule
\end{tabular}
}
\end{table}
\noindent\textbf{Comparison with SOTA on FCVID and Mini-Kinetics.}
We further evaluate the performance of our method on two large-scale video recognition benchmarks, \emph{i.e.,} FCVID and Mini-Kinetics, with ResNet-50 as the recognizer in \tabref{tab:sota_fcvid_minik}. We have a similar observation that our NSNet can achieve superior mAP with the much lower computational cost, which demonstrate the efficacy of non-saliency suppression (NS) mechanism in both untrimmed video and trimmed video scenarios.

\noindent\textbf{Practical Efficiency.}
\begin{table}[t]
\setlength{\tabcolsep}{4pt}
\renewcommand{\arraystretch}{0.9}
\scalebox{0.8}{
\begin{tabular}{ccccc}
\toprule
Method        & \multicolumn{1}{l}{mAP(\%)} & \multicolumn{1}{l}{GFLOPs} & \begin{tabular}[c]{@{}c@{}}Latency\\ \scriptsize{(bs=1)}\end{tabular} & \begin{tabular}[c]{@{}c@{}}Throughput\\ \scriptsize{(bs=32)}\end{tabular} \\
\midrule
AdaFocus \cite{adafocus}      & 75.0    & 26.6     & 181.8ms & 73.8 vid/s \\
FrameExit \cite{frameexit}     & 76.1  & 26.1     & 102.0ms               & -      \\
\textbf{Ours} & \textbf{76.8}           & \textbf{26.1}              & \textbf{42.0ms}      & \textbf{132.5 vid/s}    \\
\bottomrule
\end{tabular}
}
\caption{Comparison of practical efficiency between SotA methods. }\label{tab:speed}
\end{table}
We present the comparison results on inference speed between our NSNet and two SotA methods, FrameExit \cite{frameexit} and AdaFocus \cite{adafocus}, in \tabref{tab:speed}.  Latency and throughput with batch size of 1 and 32 are reported\footnotemark[1]. 
\footnotetext[1]{The latency and throughput results of two SotA methods are obtained by running their official code \cite{frameexit,adafocus} on the same hardware (a NVIDIA 3090 GPU with a Intel Xeon E5-2650 v3 @ 2.30GHz CPU) as ours.}   
It can be observed that our method achieves significantly superior latency (42.0 ms) and than two methods (2.4$\times$ than FrameExit \cite{frameexit} and 4.3$\times$ than AdaFocus \cite{adafocus}), which demonstrate the superiority of our parallel temporal sampling framework over existing methods on practical efficiency.

\begin{table*}[t]
\caption{{Ablation studies} on ActivityNet with mAP (\%) as the evaluation metric. Unless otherwise specified, MobileNetv2 and ResNet-101 are used as the backbone for observation network and recognizer respectively. 
}\label{tab:ablation}
\centering
\begin{subtable}[th]{0.45\textwidth}
\centering
\caption{Evaluation of the effectiveness of NS mechanism.}
\label{tab:ablation:non_saliency_modelling}
\setlength{\tabcolsep}{1.5pt}
\renewcommand{\arraystretch}{0.9}
\scalebox{0.75}{
\begin{tabular}{ccc}
\toprule
Method & VGM & FSM \\ \midrule
baseline & 73.4 & 68.4 \\
+ \textit{non-salient} class & 73.4 & 68.6  \\
+ NS & \textbf{73.8} & \textbf{74.7} \\
\bottomrule
\end{tabular}
}
\end{subtable}
\hfill
\begin{subtable}[th]{0.45\linewidth}
\caption{Performance of different number of sampled frames.}
\label{tab:ablation:num_frames}
\setlength{\tabcolsep}{4.0pt}
\centering
\scalebox{0.75}{
\renewcommand{\arraystretch}{1.2}
\begin{tabular}{cccc}
\toprule
\#F & VGM & FSM & NSNet \\ \midrule
5        & 72.6   &73.9  &\textbf{74.9}       \\
10       & 73.8   &74.7  &\textbf{75.5}       \\
\bottomrule
\end{tabular}
}
\end{subtable}
\vskip 2mm
\begin{subtable}[th]{0.45\linewidth}
\caption{Ablation of transformer encoder in feature embedding module.}
\label{tab:ablation:feature_embedding}
\centering
\setlength{\tabcolsep}{6.0pt}
\renewcommand{\arraystretch}{1.15}
\scalebox{0.75}{
    \begin{tabular}{cc}
    \toprule
    Network &  mAP(\%)      \\ \midrule
    1D Conv&   73.6   \\    
    LSTM  &   74.0   \\
	MLP   &   74.3   \\    
    Transformer &   \textbf{75.5} \\\bottomrule
    \end{tabular}
}
\end{subtable}
\hfill
\begin{subtable}[th]{0.45\textwidth}
\centering
\caption{Results of FS module with different learning objectives.}
\label{tab:ablation:objective}
\renewcommand{\arraystretch}{0.95}
\setlength{\tabcolsep}{6.0pt}
\scalebox{0.75}{
\begin{tabular}{cc}
\toprule
 FS Objective & mAP(\%)   \\ \midrule
 Baseline & 68.4 \\
 Regression & 72.0 \\
	Ranking & 72.2 \\
    Baseline+ & 73.7 \\
	Ours & \textbf{74.7} \\
	\bottomrule
\end{tabular}
}
\end{subtable}
\end{table*}

\subsection{Ablation Studies}
To comprehensively evaluate our NSNet, we provide extensive ablation studies on ActivityNet in \tabref{tab:ablation}. 
Accordingly, the effectiveness of each component in our framework is analyzed as follows. We use ResNet-101 without TSN style training as the recognizer, as the same as in \tabref{tab:simple_baseline} and \figref{fig:anet_res101} in \secref{main_results}.

\noindent\textbf{Effectiveness of non-saliency suppression.} 
We explore the effectiveness of \emph{non-saliency suppression} (NS) mechanism for two modules. For VGM, ``baseline'' denotes the variant without $\mathcal{L}_{ns}$. For FSM, ``baseline'' denotes the variant replacing our NS frame label with video label.  As shown in \tabref{tab:ablation:non_saliency_modelling}, for the VGM, simply adding a \textit{non-salient} class (from $C$ classes to $C+1$ classes) without according supervisions can not elevate performance. In contrast, by applying NS mechanism, the mAP significantly improves by 0.4\%. In the FSM, we observe that ``baseline'' present very low performance, similar to \textsc{Uniform} baseline(68.4 \vs 68.6), which is because it imposes many label noises when assigning the label of video to irrelevant or low-quality frames. We also observe that simply adding the \emph{non-salient} class cannot address the issue (68.6 \vs 68.4). NS mechanism elevates the performance significantly with effective and reasonable supervisions (74.7 \vs 68.4). More ablations in supervision signals of FSM are in \tabref{tab:ablation:objective}.


\noindent\textbf{Ablations of transformer encoder in feature embedding module.} 
\tabref{tab:ablation:feature_embedding} presents the results of different choice in FEM to passing message among the features from the input frames, including Long short-term memory networks (LSTM) \cite{lstm}, 1-D convolutional networks (1D Conv), multi-layer perceptron (MLP)~\cite{mlp}, Transformer Encoder (Transformer) \cite{transformer}.
Among all these choices, the transformer encoder achieves the highest performance.

\noindent\textbf{Different learning objectives of the FSM.}
In \tabref{tab:ablation:objective} we compare various objectives of FSM mentioned in \secref{intro} and \secref{relatedwork}, including frame classification with video labels (``baseline'', as the same one as in \tabref{tab:ablation:non_saliency_modelling}), ranking~\cite{scsampler19}, and regression~\cite{smart2020} with guiding saliency scores. 
With guiding saliency scores as supervisions, ``regression'' and ``ranking'' model saliency sampling as saliency score regression and ranking tasks respectively. They can achieve higher performance than ``baseline'' (72.0\% \& 72.2\%), whereas they overlook the exploitation of class-specific information, which limits their performance. We modify ``baseline'' by transforming hard one-hot video label to soft one-hot label using the guiding saliency score, which is denoted as ``baseline+''. This modification improves the result significantly (\textbf{73.7\%} \vs 68.4\%) by taking into account both the discrimination between salient frames and non-salient frames and the use of category-specific information. With the same setting of guiding saliency score, our NS mechanism based objective outperforms `baseline+' in a large margin (\textbf{74.7\%} \vs 73.7\%), which verifies the proposed supervisions offer more robust saliency for supplying negative samples. 

\noindent\textbf{Different numbers of sampled frames.}
As shown in \tabref{tab:ablation:num_frames}, we report the performance of different numbers of sampled frames for NSNet.
We can see that the fusion of two modules always improves the performance by 0.8\% and 1.0\% when sampling 10 and 5 frames, respectively. It demonstrates the effectiveness of our fusion strategies, especially when fewer frames are sampled. 
Besides, when fewer frames are sampled, the performance of FSM degrades more slowly than that of VGM (0.8\% \vs 1.2\%), which is because FSM can distinguish between salient frames and non-salient frames in finer granularity with the help of frame-level supervisions.


\begin{figure*}[t]
      \centering \includegraphics[width=1.0\textwidth]{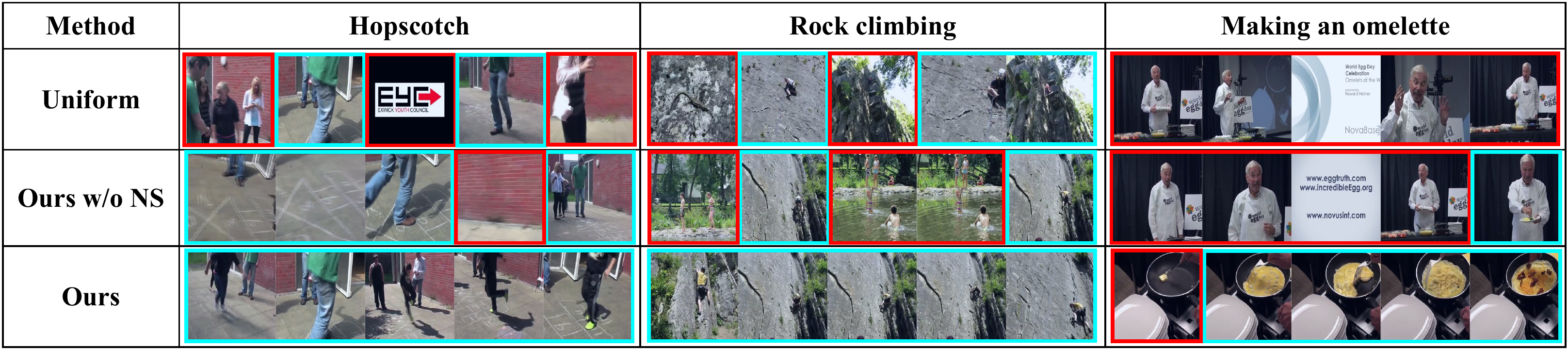}
      \caption{\textbf{Visualization of selected frames with different variants of our approach.}  \engordnumber{1} row: 
      Uniform, 
      \engordnumber{2} row: 
      Our approach without NS mechanism (Ours w/o NS),
      \engordnumber{3} row:
      Our approach (Ours). Intuitively salient frames are are outlined in \textcolor{Aqua}{aqua} while non-salient ones are outlined in \textcolor{red}{red}. Please zoom in for best view.
    }
      \label{fig:qualitative}
\end{figure*} 
\subsection{Qualitative Analysis}
\figref{fig:qualitative} shows frames sampled by different methods. Our NSNet can sample more discriminative salient frames than uniform baseline and the variant without non-saliency suppression. For example, in the \engordnumber{4} column, the \engordnumber{3} row of this column shows that ``ours w/o NS'' is mainly attracted by frames with scenes of a cook, which is not discriminative for frequently appearing in other cooking events. In contrast, \engordnumber{4} row shows NSNet can sample more indicative frames.
\section{Conclusions}
In this paper, we present the Non-saliency Suppression Network (NSNet) to measure the saliency of frames by leveraging both video-level and frame-level supervisions.
In Frame Scrutinize module, we propose a pseudo label generation strategy to enable negative sample aware frame-level saliency learning. Meanwhile, in Video Glimpse module, an attention module constrained by dual classification objectives is presented to compensate high-level information. 
Experiments show that our NSNet outperforms the state-of-the-arts on accuracy-efficiency trade-off. 
In the future, we plan to explore non-saliency suppression on both spatial and temporal dimensions to save more redundancy.

\paragraph{Acknowledgment.} Wanli Ouyang was supported by the Australian Research Council Grant DP200103223, Australian Medical Research Future Fund MRFAI000085,  CRC-P Smart Material Recovery Facility (SMRF) – Curby Soft Plastics, and CRC-P ARIA - Bionic Visual-Spatial Prosthesis for the Blind.

\clearpage
\appendix
\section*{Appendix} \label{appendix}
\setcounter{table}{0}
In this appendix, we provide more implementation details and experimental results of our proposed NSNet. Accordingly, we organize the appendix as follows.

\begin{itemize}
    \item In Section \ref{appendix:imp}, we present more implementation details for training and inference of our method. 
    \item In Section \ref{appendix:ablation}, we provide more ablation studies to further analyze the capability of our proposed NSNet. 
    \item In Section \ref{appendix:qualitative}, we present predicted saliency score distributions to qualitatively analyze the capability of proposed NSNet.
\end{itemize}

\section{Implementation Details}\label{appendix:imp}
\subsection{Training}\label{appendix:train_details}
\subsubsection{Pre-processing.}
Following previous works, the frames fed to recognizer are rescaled with a shorter side of 256 and center cropped to $224\times224$ for all datasets. The resolution of frames fed into Feature Embedding module is $224\times224$ for ActivityNet, FCVID and UCF101, and $112\times112$ for Mini-Kinetics. Note that we \textbf{only} use the RGB frames of these datasets for experiments.
Following \cite{marl}, before adaptive sampling by samplers, $T$ frames are uniformly pre-sampled from frame sequence. For those videos whose lengths are shorter than $T$, we repeat multiple times and splice them to $T$ frames. 

\subsubsection{Model training details and hyper-parameters.}
For transformer encoder, the hidden dimensions of query, key and value is set to the ratio between the number of input feature channels and the number of heads. The hidden dimension of FFN is set to be equal to the input feature channel number. Dropout~\cite{dropout} is used to reduce over-fitting. In Video Glimpse module, dropout layers are placed before classification fully-connected layer with ratio of 0.9 and after temporal attention layer with 0.2, respectively. In transformer encoder, the dropout rate after the positional encoding layer is set to 0.2. Temporal random shift is adopted as data augmentation strategy. The model is trained using SGD optimizer with momentum of 0.9 and batch size of 64 for 120 epochs. The learning rate is set to starting at $10^{-2}$, decaying by the factor of 0.1 at the \engordnumber{50} and \engordnumber{75} epoch. 

\subsubsection{Prototype generation.}
For a video $x_v$, we first apply the recognizer for each frame and obtain the predictions $\{\hat{y}_i\in\mathbb{R}^C\}_{i=1}^T$ of all frames. Then we collect the correctly predicted frames set from each video $X_g=\{x_i |_{i\in [1,T],\underset{j}{\operatorname{argmax}} \hat{y}_{i,j}=c}\}$, where $c$ is the ground truth category. We further select the top $\epsilon$ percent frames with highest confidence on the $c$-th category $\hat{y}_{i,c}$ from $X_g$ and average pool the frame features of them as the guiding video feature $\Tilde{x}^g_v$. Then, for the $c$-th category, the prototype feature $p_c$ can be computed by average pooling the all the guiding video features belonging to the $c$-th category. We use  $\epsilon=30$ in all experiments.

\subsection{Inference} \label{appendix:test_details}
We describe the combination strategies in detail here.

\subsection{Score Combination.}
We consider 3 types of fusion operations, which includes \textit{addition}, \textit{multiplication} and \textit{maximization}. For \textit{addition}, we fuse the saliency scores of two branches in convex combination $\alpha {s}^{f}_{i} + (1-\alpha){s}^{v}_{i} $, where $\alpha$ is a combination ratio parameter. For \textit{multiplication} and \textit{maximization}, we fuse the saliency scores of two branches in element-wise muliplication ${s}^{f}_{i} * {s}^{v}_{i}$ and element-wise maximization $\max({s}^{f}_{i}, {s}^{v}_{i})$, respectively. 

\subsubsection{Index Combination.}
We consider three strategies, which involves \textit{intersection}, \textit{union} and \textit{join}. We firstly get frame index lists $\{{\pi}_i^{f}\}_{i=1}^T$ and $\{{\pi}_i^{v}\}_{i=1}^T$ by sorting $\{{s}^{f}_{i}\}_{i=1}^T$ and $\{{s}^{v}_{i}\}_{i=1}^T$ in descending order, respectively. In \textit{intersection}, given a budget of $K$ salient frames at most, we firstly take top $K$ frames from index lists, $\{{\pi}_{i}^{f}\}_{i=1}^K$ and $\{{\pi}_{i}^{v}\}_{i=1}^K$ respectively and get the intersection of them $I(K) = \{{\pi}_{i}^{f}\}_{i=1}^K \cap \{{\pi}_{i}^{f}\}_{i=1}^K$. When there exist coincident frames, we expand $I$ with one element from either $\{{\pi}_{i}^{f}\}_{i=K+1}^T$ or $\{{\pi}_{i}^{v}\}_{i=K+1}^T$ by turns for $i^{\prime}$ steps, until $|I(K+i^{\prime})|=K$. For \textit{union}, following \cite{scsampler19}, we try to obtain a set of salient frames whose length is represented by $\alpha|{\pi}^{f}|+(1-\alpha)|{\pi}^{v}|$. We firstly get the union of top saliency frames from two lists $U(K) = \{{\pi}^{f}_{i}\}_{i=1}^{\lceil K*\alpha \rceil} \cup \{{\pi}_{i}^{v}\}_{i=1}^{\lceil K*(1-\alpha) \rceil}$. We expand $U(K)$ with one element from ${\pi}^{f}$ at a time for $i^{\prime}$ steps until $|U(K+i^{\prime})|=K$. For \textit{join}, we concatenate $\{s_i^{f}\}_{i=1}^T$ and $\{s_i^{v}\}_{i=1}^T$ to a list with a length of $2T$, from which $K$ non-overlap top saliency score frames are selected as final salient frames set.

We use $\alpha=0.6$ for score addition and index union in the ablation studies of fusion strategies. Union fusion are used in all other experiments. 

\section{Additional Ablation Studies} \label{appendix:ablation}

\subsection{Different guiding saliency score.}
In \tabref{tab:ablation:guiding_saliency_score}, we compare our prototype based guiding saliency score with an alternative choice, where we use the classification response of the ground truth category produced by the recognizer to generate the NS pseudo labels, namely \emph{response-based} guiding saliency score. It is shown that the prototype based score achieves better performance than response based one, which demonstrates that the prototype distance in feature space can offer more robust saliency cues.
\begin{table}[h]
\centering
\caption{Performance of different guiding saliency score in FS module.}
\renewcommand{\arraystretch}{1.15}
\scalebox{1.0}{
\begin{tabular}{cc}
\toprule
Guiding Saliency Score & \multicolumn{1}{l}{mAP(\%)} \\ \midrule
Response-based & 74.1 \\
Prototype-based & \textbf{74.7} \\
\bottomrule
\end{tabular}
}
\label{tab:ablation:guiding_saliency_score}
\end{table} 
\subsection{Different fusion strategies of two modules.}
In \tabref{tab:ablation:combination_strategy}, we show the impacts of different fusion strategies which are described in \secref{appendix:test_details}. We can observe that various fusion strategies consistently improve the performance of single modules. The `index union' fusion gets slightly higher performance than others thus we choose it in all our experiments.

\begin{table}[h]
\centering
\caption{Comparison of various fusion strategies.}
\label{tab:ablation:combination_strategy}
\setlength{\tabcolsep}{4.5pt}
\renewcommand{\arraystretch}{0.9}
\scalebox{1.0}{
\begin{tabular}{cccc}
\toprule
& Max   & Mul  & Add  \\ 
\multicolumn{1}{c}{Score} & 75.1 & 75.2 & \textbf{75.3} \\\midrule
& Join & Inter & Union   \\
\multicolumn{1}{c}{Index} & 75.1 & 74.9 & \textbf{75.5}\\
\bottomrule
\end{tabular}
}
\end{table}
\subsection{Different lightweight Feature Extractor in FEM.}
In \tabref{tab:ablation:observation_network} we compare various backbones for lightweight feature extractor in FEM. 
As expected, the lightweight backbone with better performance is complementary to our method.
Comparing with the ShuffleNetv2 \cite{shufflenetv2} and MobileNetv2~\cite{mobilenetv2} counterparts, our NSNet gets additional improvement on EfficientNet-b0 \cite{efficientnet} with extra computation overhead.
For fair comparisons with previous works, we use the MobileNetv2 as the lightweight feature extractor by default.
\begin{table}[h]
\centering
\scalebox{1.0}{
    \begin{tabular}{ccc}
    \toprule
    Backbone & mAP & FLOPs/f \\ \midrule
    ShuffleNetv2 &  70.8 & 0.15G \\
    MobileNetv2  &  75.5 & 0.31G \\
    EfficientNet-b0  &  76.0 & 0.39G \\
    \bottomrule
    \end{tabular}
}
\caption{Study on different backbones for lightweight extractor in FEM. FLOPs/f means FLOPs for each frame processed by the backbone.}
\label{tab:ablation:observation_network}
\end{table}
\subsection{Different recognizer.}
In \tabref{tab:ablation:recognizer} we investigate the impacts of various backbones of the recognizer, where we also report the training strategy of recognizer, \emph{i.e.,} with TSN or without TSN. The model without TSN is trained by sampling one frame from each video. 
It is shown that our method is complementary to more advanced recognizers.
\begin{table}[h]
\centering
\scalebox{1.0}{
    \begin{tabular}{ccc}
    \toprule
    Backbone & Train & mAP(\%) \\ \midrule
    ResNet-101 &  w/o TSN & 75.5 \\
    ResNet-101  &  w/ TSN & 80.8 \\
    ResNet-152  &  w/ TSN & 83.0 \\
    \bottomrule
    \end{tabular}
}
\caption{Study on different backbones for the recognizer. ``Train'' refers to training strategy, \emph{viz.,} with TSN style training or without TSN style Training.}
\label{tab:ablation:recognizer}
\end{table}


\section{Qualitative Analysis} \label{appendix:qualitative}
\begin{figure*}[t!]
      \centering \includegraphics[width=1.0\textwidth]{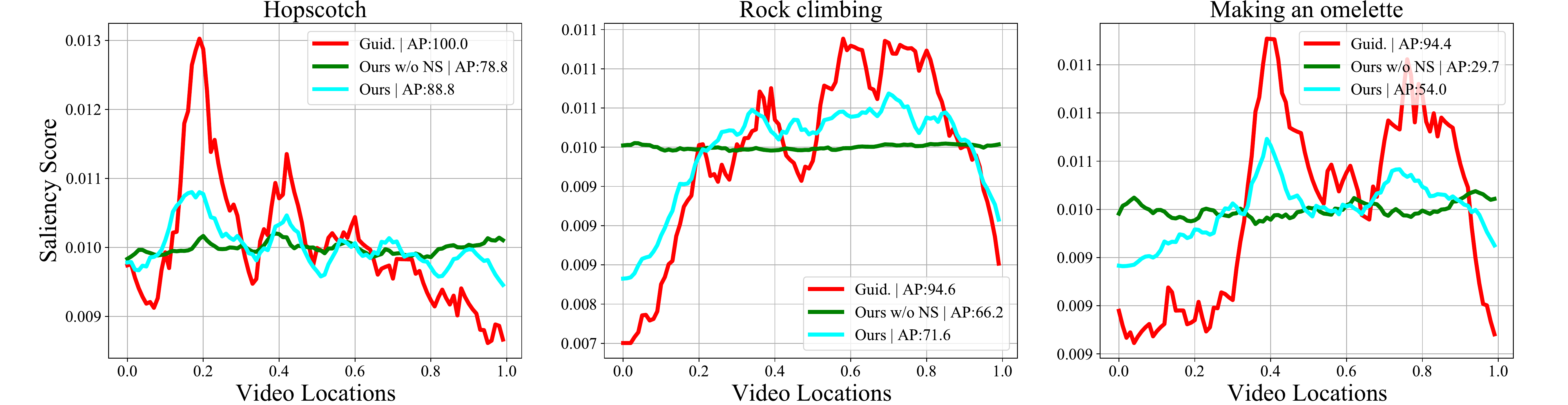}
      \caption{\textbf{Measured saliency distribution by different variants of our approach.} We show the average value over all samples for a category for 3 variants, \emph{viz.,} guiding saliency score (Guid.), our approach without non-saliency suppression (Ours w/o NS) and our approach(Ours). Our approach can generate saliency measurements close to Guid. for all 3 categories. However, Ours w/o NS only produce a relatively flat line on difficult categories like \emph{rock climbing} and \emph{making an omelette}, which shows it cannot handle saliency measurement on difficult categories without NS mechansim. Predicted saliency distributions are smoothed by Exponential Moving Average with weight of 0.8 for a better sense of trend.}  
      \label{fig:sal_curve}
\end{figure*}
To take a closer look to how NS mechanism benefits saliency measurement, we present temporal saliency distributions produced by variants of our approach on the validation set of ActivityNet in \figref{fig:sal_curve}, which is computed by averaging the temporal saliency distribution of all samples within a given class. We adopt guiding saliency score (Guid.) as an alternative of saliency ``ground truth'', for it exploits labels of validation set and represents a upper bound of any sampler, achieving mAP of 96.3 (\emph{v.s.} 75.3 achieved by NSNet). 
We can see that the saliency distribution of NSNet is more close to that of Guid. and achieving much higher AP than that of NSNet w/o NS on all three categories. As the discrimination difficulty increases, AP decreases dramatically from left sub-figures to right ones in \figref{fig:sal_curve}. In the easiest category \emph{hopscotch}, both NSNet and NSNet w/o NS show similar saliency trends to Guid. to varying degrees. However, in much more difficult categories with low AP, like \emph{rocking climbing} and \emph{making a omelette}, the saliency scores measured by NSNet w/o NS tend to generate temporal uniform distributions and NSNet still shows highly similar trends to Guid., which demonstrates that proposed NS based supervisions can enhance robustness of saliency measurements in many scenarios.

%
%
\bibliographystyle{splncs04}
\bibliography{egbib}

\begin{thebibliography}{10}
\providecommand{\url}[1]{\texttt{#1}}
\providecommand{\urlprefix}{URL }
\providecommand{\doi}[1]{https://doi.org/#1}

\bibitem{timesformer}
Bertasius, G., Wang, H., Torresani, L.: Is space-time attention all you need
  for video understanding? arXiv preprint arXiv:2102.05095  (2021)

\bibitem{caba2015activitynet}
Caba~Heilbron, F., Escorcia, V., Ghanem, B., Carlos~Niebles, J.: Activitynet: A
  large-scale video benchmark for human activity understanding. In: Proceedings
  of the ieee conference on computer vision and pattern recognition. pp.
  961--970 (2015)

\bibitem{i3d}
Carreira, J., Zisserman, A.: Quo vadis, action recognition? a new model and the
  kinetics dataset. In: proceedings of the IEEE Conference on Computer Vision
  and Pattern Recognition. pp. 6299--6308 (2017)

\bibitem{AKnet}
Chen, X., Han, Y., Wang, X., Sun, Y., Yang, Y.: Action keypoint network for
  efficient video recognition. arXiv preprint arXiv:2201.06304  (2022)

\bibitem{fastforward}
Fan, H., Xu, Z., Zhu, L., Yan, C., Ge, J., Yang, Y.: Watching a small portion
  could be as good as watching all: Towards efficient video classification. In:
  IJCAI International Joint Conference on Artificial Intelligence (2018)

\bibitem{mamico}
Fang, B., Wu, W., Liu, C., Zhou, Y., He, D., Wang, W.: Mamico: Macro-to-micro
  semantic correspondence for self-supervised video representation learning. In
  Proc. ACMMM  (2022)

\bibitem{slowfast}
Feichtenhofer, C., Fan, H., Malik, J., He, K.: Slowfast networks for video
  recognition. In: Proceedings of the IEEE/CVF international conference on
  computer vision. pp. 6202--6211 (2019)

\bibitem{listentolook}
Gao, R., Oh, T.H., Grauman, K., Torresani, L.: Listen to look: Action
  recognition by previewing audio. In: Proceedings of the IEEE/CVF Conference
  on Computer Vision and Pattern Recognition. pp. 10457--10467 (2020)

\bibitem{frameexit}
Ghodrati, A., Bejnordi, B.E., Habibian, A.: Frameexit: Conditional early
  exiting for efficient video recognition. In: Proceedings of the IEEE/CVF
  Conference on Computer Vision and Pattern Recognition. pp. 15608--15618
  (2021)

\bibitem{smart2020}
Gowda, S.N., Rohrbach, M., Sevilla-Lara, L.: {SMART} frame selection for action
  recognition  \textbf{35}(2),  1451--1459 (2021),
  \url{https://ojs.aaai.org/index.php/AAAI/article/view/16235}

\bibitem{dynamic_survey}
Han, Y., Huang, G., Song, S., Yang, L., Wang, H., Wang, Y.: Dynamic neural
  networks: A survey. IEEE Transactions on Pattern Analysis and Machine
  Intelligence  (2021)

\bibitem{resnet}
He, K., Zhang, X., Ren, S., Sun, J.: Deep residual learning for image
  recognition. In: Proceedings of the IEEE conference on computer vision and
  pattern recognition. pp. 770--778 (2016)

\bibitem{lstm}
Hochreiter, S., Schmidhuber, J.: Long short-term memory. Neural computation
  \textbf{9}(8),  1735--1780 (1997)

\bibitem{ASCNet}
Huang, D., Wu, W., Hu, W., Liu, X., He, D., Wu, Z., Wu, X., Tan, M., Ding, E.:
  Ascnet: Self-supervised video representation learning with appearance-speed
  consistency. In: Proceedings of the IEEE/CVF International Conference on
  Computer Vision. pp. 8096--8105 (2021)

\bibitem{huang2018toward}
Huang, W., Fan, L., Harandi, M., Ma, L., Liu, H., Liu, W., Gan, C.: Toward
  efficient action recognition: Principal backpropagation for training
  two-stream networks. IEEE Transactions on Image Processing  \textbf{28}(4),
  1773--1782 (2018)

\bibitem{haoran_ijcai}
Ji, Z., Chen, K., Wang, H.: Step-wise hierarchical alignment network for
  image-text matching. In: Zhou, Z. (ed.) Proceedings of the Thirtieth
  International Joint Conference on Artificial Intelligence, {IJCAI} 2021,
  Virtual Event / Montreal, Canada, 19-27 August 2021. pp. 765--771. ijcai.org
  (2021). \doi{10.24963/ijcai.2021/106},
  \url{https://doi.org/10.24963/ijcai.2021/106}

\bibitem{fcvid}
Jiang, Y.G., Wu, Z., Wang, J., Xue, X., Chang, S.F.: Exploiting feature and
  class relationships in video categorization with regularized deep neural
  networks. IEEE Transactions on Pattern Analysis and Machine Intelligence
  \textbf{40}(2),  352--364 (2018). \doi{10.1109/TPAMI.2017.2670560}

\bibitem{kay2017kinetics}
Kay, W., Carreira, J., Simonyan, K., Zhang, B., Hillier, C., Vijayanarasimhan,
  S., Viola, F., Green, T., Back, T., Natsev, P., et~al.: The kinetics human
  action video dataset. arXiv preprint arXiv:1705.06950  (2017)

\bibitem{dynamicSTE}
Kim, H., Jain, M., Lee, J.T., Yun, S., Porikli, F.: Efficient action
  recognition via dynamic knowledge propagation. In: Proceedings of the
  IEEE/CVF International Conference on Computer Vision. pp. 13719--13728 (2021)

\bibitem{scsampler19}
Korbar, B., Tran, D., Torresani, L.: Scsampler: Sampling salient clips from
  video for efficient action recognition. In: Proceedings of the IEEE/CVF
  International Conference on Computer Vision (ICCV) (October 2019)

\bibitem{ada3d}
Li, H., Wu, Z., Shrivastava, A., Davis, L.S.: 2d or not 2d? adaptive 3d
  convolution selection for efficient video recognition. In: Proceedings of the
  IEEE/CVF Conference on Computer Vision and Pattern Recognition. pp.
  6155--6164 (2021)

\bibitem{tea2020}
Li, Y., Ji, B., Shi, X., Zhang, J., Kang, B., Wang, L.: Tea: Temporal
  excitation and aggregation for action recognition. In: CVPR. pp. 909--918
  (2020)

\bibitem{tsm}
Lin, J., Gan, C., Han, S.: Tsm: Temporal shift module for efficient video
  understanding. In: Proceedings of the IEEE/CVF International Conference on
  Computer Vision. pp. 7083--7093 (2019)

\bibitem{ocsampler}
Lin, J., Duan, H., Chen, K., Lin, D., Wang, L.: Ocsampler: Compressing videos
  to one clip with single-step sampling. In: Proceedings of the IEEE/CVF
  Conference on Computer Vision and Pattern Recognition. pp. 13894--13903
  (2022)

\bibitem{liu2019multi}
Liu, Y., Ma, L., Zhang, Y., Liu, W., Chang, S.F.: Multi-granularity generator
  for temporal action proposal. In: Proceedings of the IEEE/CVF conference on
  computer vision and pattern recognition. pp. 3604--3613 (2019)

\bibitem{swintransformer}
Liu, Z., Lin, Y., Cao, Y., Hu, H., Wei, Y., Zhang, Z., Lin, S., Guo, B.: Swin
  transformer: Hierarchical vision transformer using shifted windows. In:
  Proceedings of the IEEE/CVF International Conference on Computer Vision. pp.
  10012--10022 (2021)

\bibitem{arnet}
Meng, Y., Lin, C.C., Panda, R., Sattigeri, P., Karlinsky, L., Oliva, A.,
  Saenko, K., Feris, R.: Ar-net: Adaptive frame resolution for efficient action
  recognition. In: European Conference on Computer Vision. pp. 86--104.
  Springer (2020)

\bibitem{adafuse}
Meng, Y., Panda, R., Lin, C.C., Sattigeri, P., Karlinsky, L., Saenko, K.,
  Oliva, A., Feris, R.: Adafuse: Adaptive temporal fusion network for efficient
  action recognition. arXiv preprint arXiv:2102.05775  (2021)

\bibitem{bgmodel}
Nguyen, P.X., Ramanan, D., Fowlkes, C.C.: Weakly-supervised action localization
  with background modeling. In: Proceedings of the IEEE/CVF International
  Conference on Computer Vision. pp. 5502--5511 (2019)

\bibitem{adamml}
Panda, R., Chen, C.F., Fan, Q., Sun, X., Saenko, K., Oliva, A., Feris, R.:
  Adamml: Adaptive multi-modal learning for efficient video recognition. arXiv
  preprint arXiv:2105.05165  (2021)

\bibitem{p3d}
Qiu, Z., Yao, T., Mei, T.: Learning spatio-temporal representation with
  pseudo-3d residual networks. In: proceedings of the IEEE International
  Conference on Computer Vision. pp. 5533--5541 (2017)

\bibitem{mobilenetv2}
Sandler, M., Howard, A., Zhu, M., Zhmoginov, A., Chen, L.C.: Mobilenetv2:
  Inverted residuals and linear bottlenecks. In: Proceedings of the IEEE
  conference on computer vision and pattern recognition. pp. 4510--4520 (2018)

\bibitem{proto_network}
Snell, J., Swersky, K., Zemel, R.: Prototypical networks for few-shot learning.
  Advances in neural information processing systems  \textbf{30} (2017)

\bibitem{ucf101}
Soomro, K., Zamir, A.R., Shah, M.: Ucf101: A dataset of 101 human actions
  classes from videos in the wild. arXiv preprint arXiv:1212.0402  (2012)

\bibitem{surui_2019_CVPR}
Su, R., Ouyang, W., Zhou, L., Xu, D.: Improving action localization by
  progressive cross-stream cooperation. In: Proceedings of the IEEE/CVF
  Conference on Computer Vision and Pattern Recognition (CVPR) (June 2019)

\bibitem{surui_2021_ICCV}
Su, R., Yu, Q., Xu, D.: Stvgbert: A visual-linguistic transformer based
  framework for spatio-temporal video grounding. In: Proceedings of the
  IEEE/CVF International Conference on Computer Vision (ICCV). pp. 1533--1542
  (October 2021)

\bibitem{videoiq}
Sun, X., Panda, R., Chen, C.F.R., Oliva, A., Feris, R., Saenko, K.: Dynamic
  network quantization for efficient video inference. In: Proceedings of the
  IEEE/CVF International Conference on Computer Vision. pp. 7375--7385 (2021)

\bibitem{labelsmooth}
Szegedy, C., Vanhoucke, V., Ioffe, S., Shlens, J., Wojna, Z.: Rethinking the
  inception architecture for computer vision. In: Proceedings of the IEEE
  conference on computer vision and pattern recognition. pp. 2818--2826 (2016)

\bibitem{wsod_oicr}
Tang, P., Wang, X., Bai, X., Liu, W.: Multiple instance detection network with
  online instance classifier refinement. In: Proceedings of the IEEE conference
  on computer vision and pattern recognition. pp. 2843--2851 (2017)

\bibitem{r2plus1d}
Tran, D., Wang, H., Torresani, L., Ray, J., LeCun, Y., Paluri, M.: A closer
  look at spatiotemporal convolutions for action recognition. In: CVPR (2018)

\bibitem{transformer}
Vaswani, A., Shazeer, N., Parmar, N., Uszkoreit, J., Jones, L., Gomez, A.N.,
  Kaiser, {\L}., Polosukhin, I.: Attention is all you need. In: Advances in
  neural information processing systems. pp. 5998--6008 (2017)

\bibitem{tsn}
Wang, L., Xiong, Y., Wang, Z., Qiao, Y., Lin, D., Tang, X., Van~Gool, L.:
  Temporal segment networks: Towards good practices for deep action
  recognition. In: European conference on computer vision. pp. 20--36. Springer
  (2016)

\bibitem{wang2020symbiotic}
Wang, X., Zhu, L., Wu, Y., Yang, Y.: Symbiotic attention for egocentric action
  recognition with object-centric alignment. IEEE transactions on pattern
  analysis and machine intelligence  (2020)

\bibitem{adafocus}
Wang, Y., Chen, Z., Jiang, H., Song, S., Han, Y., Huang, G.: Adaptive focus for
  efficient video recognition. arXiv preprint arXiv:2105.03245  (2021)

\bibitem{gfnet}
Wang, Y., Lv, K., Huang, R., Song, S., Yang, L., Huang, G.: Glance and focus: a
  dynamic approach to reducing spatial redundancy in image classification.
  Advances in Neural Information Processing Systems  \textbf{33},  2432--2444
  (2020)

\bibitem{adafocusv2}
Wang, Y., Yue, Y., Lin, Y., Jiang, H., Lai, Z., Kulikov, V., Orlov, N., Shi,
  H., Huang, G.: Adafocus v2: End-to-end training of spatial dynamic networks
  for video recognition. In: Proceedings of the IEEE/CVF Conference on Computer
  Vision and Pattern Recognition (CVPR). pp. 20062--20072 (June 2022)

\bibitem{wsss}
Wei, Y., Feng, J., Liang, X., Cheng, M.M., Zhao, Y., Yan, S.: Object region
  mining with adversarial erasing: A simple classification to semantic
  segmentation approach. In: Proceedings of the IEEE conference on computer
  vision and pattern recognition. pp. 1568--1576 (2017)

\bibitem{mlp}
Werbos, P.J.: Applications of advances in nonlinear sensitivity analysis. In:
  System modeling and optimization, pp. 762--770. Springer (1982)

\bibitem{wu2021weakly}
Wu, J., Zhang, W., Li, G., Wu, W., Tan, X., Li, Y., Ding, E., Lin, L.:
  Weakly-supervised spatio-temporal anomaly detection in surveillance video.
  IJCAI  (2021)

\bibitem{mvf}
Wu, W., He, D., Lin, T., Li, F., Gan, C., Ding, E.: Mvfnet: Multi-view fusion
  network for efficient video recognition. In: Proceedings of the AAAI
  Conference on Artificial Intelligence. vol.~35, pp. 2943--2951 (2021)

\bibitem{marl}
Wu, W., He, D., Tan, X., Chen, S., Wen, S.: Multi-agent reinforcement learning
  based frame sampling for effective untrimmed video recognition. In:
  Proceedings of the IEEE/CVF International Conference on Computer Vision. pp.
  6222--6231 (2019)

\bibitem{wu2020dynamic}
Wu, W., He, D., Tan, X., Chen, S., Yang, Y., Wen, S.: Dynamic inference: A new
  approach toward efficient video action recognition. In: Proceedings of the
  IEEE/CVF Conference on Computer Vision and Pattern Recognition Workshops. pp.
  676--677 (2020)

\bibitem{Wu2022TransferringTK}
Wu, W., Sun, Z., Ouyang, W.: Transferring textual knowledge for visual
  recognition. arXiv e-prints pp. arXiv--2207 (2022)

\bibitem{dsanet}
Wu, W., Zhao, Y., Xu, Y., Tan, X., He, D., Zou, Z., Ye, J., Li, Y., Yao, M.,
  Dong, Z., et~al.: Dsanet: Dynamic segment aggregation network for video-level
  representation learning. In Proc. ACMMM  (2021)

\bibitem{liteeval}
Wu, Z., Xiong, C., Jiang, Y.G., Davis, L.S.: Liteeval: A coarse-to-fine
  framework for resource efficient video recognition. arXiv preprint
  arXiv:1912.01601  (2019)

\bibitem{adaframe}
Wu, Z., Xiong, C., Ma, C.Y., Socher, R., Davis, L.S.: Adaframe: Adaptive frame
  selection for fast video recognition. In: Proceedings of the IEEE/CVF
  Conference on Computer Vision and Pattern Recognition. pp. 1278--1287 (2019)

\bibitem{tsqnet}
Xia, B., Wang, Z., Wu, W., Wang, H., Han, J.: Temporal saliency query network
  for efficient video recognition. ECCV  (2022)

\bibitem{s3d}
Xie, S., Sun, C., Huang, J., Tu, Z., Murphy, K.: Rethinking spatiotemporal
  feature learning: Speed-accuracy trade-offs in video classification. In: ECCV
  (2018)

\bibitem{bcnet}
Yang, H., Wu, W., Wang, L., Jin, S., Xia, B., Yao, H., Huang, H.: Temporal
  action proposal generation with background constraint. In: Proceedings of the
  AAAI Conference on Artificial Intelligence. vol.~36, pp. 3054--3062 (2022)

\bibitem{frameglimpse}
Yeung, S., Russakovsky, O., Mori, G., Fei-Fei, L.: End-to-end learning of
  action detection from frame glimpses in videos. In: Proceedings of the IEEE
  conference on computer vision and pattern recognition. pp. 2678--2687 (2016)

\bibitem{zhang2020discriminability}
Zhang, M., Song, G., Zhou, H., Liu, Y.: Discriminability distillation in group
  representation learning. In: European Conference on Computer Vision. pp.
  1--19. Springer (2020)

\bibitem{dsn}
Zheng, Y.D., Liu, Z., Lu, T., Wang, L.: Dynamic sampling networks for efficient
  action recognition in videos. IEEE Transactions on Image Processing
  \textbf{29},  7970--7983 (2020)

\bibitem{rra}
Zhu, C., Tan, X., Zhou, F., Liu, X., Yue, K., Ding, E., Ma, Y.: Fine-grained
  video categorization with redundancy reduction attention. In: Proceedings of
  the European Conference on Computer Vision (ECCV). pp. 136--152 (2018)

\end{thebibliography}
\end{document}